Subject Section

# MS-ConTab: Multi-Scale Contrastive Learning of Mutation Signatures for Pan-Cancer Representation and Stratification

Yifan Dou[1], Adam Khadre[1], Ruben C. Petreaca[2,3], Mirzaei Golrokh[1*]

[1]Department of Computer Science and Engineering, Ohio State University, Columbus, OH 43210, United States
[2]Department of Molecular Genetics, Ohio State University,Marion, OH 43302, United States
[3]Cancer Biology Program, James Comprehensive Cancer Center, Ohio State University, Columbus, OH 43210,
*Corresponding Author. Email: mirzaei.4@osu.edu

## Abstract

**Motivation:** Understanding pan-cancer level mutational landscape offers critical insights into the molecular mechanisms underlying tumorigenesis. While patient-level machine learning techniques have been widely employed to identify tumor subtypes, cohort-level clustering—where entire cancer types are grouped based on shared molecular features—has largely relied on classical statistical methods.

**Results:** In this study, we introduce a novel unsupervised contrastive learning framework to cluster 43 cancer types based on coding mutation data derived from the COSMIC database. For each cancer type, we construct two complementary mutation signatures: a gene-level profile capturing nucleotide substitution patterns across the most frequently mutated genes, and a chromosome-level profile representing normalized substitution frequencies across chromosomes. These dual views are encoded using TabNet encoders and optimized via a multi-scale contrastive learning objective (NT-Xent loss) to learn unified cancer-type embeddings. We demonstrate that the resulting latent representations yield biologically meaningful clusters of cancer types, aligning with known mutational processes and tissue origins. Our work represents the first application of contrastive learning to cohort-level cancer clustering, offering a scalable and interpretable framework for mutation-driven cancer subtyping.

**Availability and Implementation:** Data and Code are available at: https://github.com/anonymous2025Aug/MS-ConTab

**Contact**: mirzaei.4@osu.edu

**Supplementary information:** Supplementary material includes **Supplementary Table 1- 3** and **Supplementary Figure 1**, which provide additional data supporting the main results.

## 1 Introduction

Cancer is a highly heterogeneous disease, driven by complex genomic alterations that vary across and within tumor types. Understanding this heterogeneity is essential for improving cancer diagnosis, prognostication, and treatment selection. Large-scale cancer genomics initiatives, such as The Cancer Genome Atlas (TCGA) (The Cancer Genome Atlas *et al.* 2013) have enabled comprehensive characterization of tumors at multiple molecular levels, including gene-level mutations (Mirzaei 2023) and chromosomal alterations and copy number variations (Mirzaei and Petreaca 2022). However, effectively integrating and representing these diverse, multi-scale genomic features for unsupervised cancer clustering remains a significant challenge.

Two complementary perspectives have emerged to support this goal: patient-level subtyping and cancer-type (cohort-level) clustering. Patient-level subtyping focuses on stratifying individual tumors into biologically meaningful subgroups based on molecular profiles, such as gene expression or mutations. This approach has led to clinically actionable subtypes in several cancers such as breast cancer (Orrantia-Borunda *et al.* 2022) and glioblastoma (Sidaway 2017), guiding precision treatment strategies. A range of molecular features have been used for this purpose, including chromosomal rearrangements (Mirzaei 2022), gene expression profiles (Al-Azani *et al.* 2024), and multi-omics data including RNA sequence, copy number variants, DNA methylation, etc. (Raj and Mirzaei 2022, Raj *et al.* 2024). In contrast, cancer-type or cohort-level clustering aims to group entire cancer types or tissue lineages based on shared genomic characteristics. Leveraging such large-scale datasets makes it possible to uncover higher-level patterns, such as conserved mutational processes and evolutionary constraints, that span across tumor types. These broader insights offer opportunities for drug repurposing, biomarker discovery, and a deeper understanding of cancer biology. Despite its promise, cohort-level clustering is still dominated by conventional methods like hierarchical clustering and non-negative matrix factorization (NMF) (Brunet *et al. 2004*), which often fall short in capturing complex, nonlinear relationships among genomic features.

Recent advances in representation learning, particularly deep learning and contrastive learning, have revolutionized patient-level subtyping by learning informative, low-dimensional embeddings from high-dimensional omics data. Yet, these methods remain underexplored for cohort-level tasks. Alexandrov *et al.* (2020) performed a landmark pan-cancer mutational signature analysis using unsupervised NMF-based methods (SigProfiler, SignatureAnalyzer) to extract substitution and indel-based mutational signatures from aggregated mutation spectra. While this work has been foundational in linking mutational processes to their biological mechanisms, its reliance on prior signature decomposition inherently compresses and smooths the data, potentially discarding fine-grained, cancer-type–specific variation. Moreover, NMF requires manual tuning of the number of latent factors and assumes a linear mixture model, which may not fully capture complex, nonlinear dependencies in mutation patterns across cancers.

In this work, we address these gaps by proposing MS-ConTab (*Multi-Scale Contrastive TabNet*), a fully unsupervised framework for cohort-level cancer clustering. MS-ConTab learns unified embeddings from gene-level and chromosome-level protein-coding single-nucleotide substitution profiles using TabNet encoders and a contrastive NT-Xent loss. By focusing on protein-coding substitutions in frequently mutated genes, we hypothesize that our multi-scale contrastive approach can capture higher-order structures across cancer types. Unlike traditional approaches, MS-ConTab does not require predefined labels or cluster counts, and it leverages multi-scale contrastive pretraining to generate robust, biologically meaningful representations for downstream clustering. Our contributions are threefold:

- Introduction of MS-ConTab, the first contrastive learning framework tailored for cohort-level cancer clustering using protein-coding nucleotide substitution signatures.
- Combination of gene-level and chromosome-level mutation views within a unified embedding space to capture hierarchical genomic structure.
- Identification of two distinct, biologically coherent cancer clusters, revealing shared mutational processes across cancer types.
- Use of TabNet encoders for feature selection, enhancing interpretability of multi-scale genomic representations.
- Rigorous benchmarking against traditional and deep learning baselines (NMF, autoencoder, SimCLR, DeepCluster, hierarchical clustering), demonstrating superior cluster quality.

# 2 Related Works

A wide spectrum of machine learning techniques has been applied to cancer subtyping, with approaches spanning unsupervised to supervised paradigms and addressing both patient-level and cohort-level analyses. Unsupervised learning methods discover latent structure without labels while supervised strategies leverage existing clinical or histopathological annotations. Moreover, studies have addressed both patient-level subtyping, where individual tumor samples are clustered, and cohort-level analyses, which group entire cancer types based on shared molecular signatures.

## 2.1 Patient-level clustering

*Unsupervised subtyping*- This approach has been widely explored using both statistical and machine leaning approaches. Statistical techniques include Non-negative Matrix Factorization (NMF) (Brunet *et al*. 2004), Multiple Canonical Correlation Analysis (MCCA) (Witten and Tibshirani 2009), iCluster (Shen *et al*. 2009), iCluster+ (Mo *et al*. 2013), and DeepCluster (Caron *et al*. 2018). Alexandrov *et al*. (2013) applied NMF to single-base substitution profiles, identifying recurrent COSMIC mutational signatures followed by hierarchical consensus clustering. More recently, deep learning models such as autoencoders and variational autoencoders (VAEs) have been employed to compress high-dimensional omics data into latent representations, which are then clustered using algorithms such as k-means or Gaussian mixture models (Way *et al*. 2018). Zhang and Kiryu (2022) integrated gene expression, methylation, and miRNA data into a unified latent manifold via optimization and minimized the Kullback–Leibler divergence to assign patient clusters, validating their findings through survival analysis and pathway enrichment. Chen *et al*. (2023) developed a multi-omic autoencoder framewoek that extracted shared and modality-specific components from mRNA expression, miRNA expression, and DNA methylation, applied contrastive learning to align the shared components, and performed clustering using k-means.

*Supervised and semi-supervised subtyping*- Supervised methods guide representation learning using known labels such as tumor stage, survival outcomes, or histological subtype. Metric learning techniques (e.g., Siamese and triplet networks) have been used to learn distance metrics that cluster samples of the same subtype. Arora et al. (2020) introduced survClust, which learns a feature-weighted distance metric across multi-omic data and applies k-means to stratify patients by survival, improving prognostic subgrouping. Random forest-based feature selection followed by clustering has also been used to reduce noise and emphasize subtype-relevant features. Dou and Mirzaei (2025) applied Graph Convolutional Networks (GCNs) and Graph Attention Networks (GATs) for supervised classification of cancer subtypes from multi-omic data (DNA methylation, RNA sequence, etc.). Although most supervised efforts are focused on classification or survival prediction, their learned representations can indirectly aid clustering, subtype refinement, and visualization of disease heterogeneity. In settings with limited labeled samples, self-supervised and semi-supervised approaches have been employed. For example, Chen *et al*. (2023) designed a self-supervised framework that learns robust representations of multi-omic profiles through contrastive pretraining, enabling more effective clustering and subtype discovery with minimal supervision.

## 2.2 Cohort-level clustering

This clustering approach operates at the level of cancer types, rather than individual tumors, to reveal broader pan-cancer relationships. Compared to patient-level subtyping, cohort analyses reduce granularity but highlight fundamental biological processes shared across tumor origins. Bailey *et al*. (2018) applied NMF to mutation frequency profiles aggregated by cancer type and used hierarchical consensus clustering on COSMIC signature exposures to group cancers by dominant mutational processes. Hoadley *et al*. (2014) integrated mRNA expression, DNA methylation, miRNA, and copy-number data across 12 TCGA cancer types, using hierarchical consensus clustering to define "pan-cancer classes" of tumor entities. Additional cohort-level studies include hierarchical clustering of tissue-specific gene expression centroids across GTEx and TCGA to map cancer–normal relationships, and self-organizing maps on aggregated proteomic profiles to classify tumor lineages.

Despite these improvements, advanced representation learning techniques remain underutilized for cohort-level clustering. Most cohort-level studies rely on classical statistical or clustering techniques such as hierarchical clustering or NMF, without leveraging learned data representations. Hierarchical clustering, in particular, is favored due to its flexibility, as it does not require re-specifying the number of clusters, unlike other techniques such as k-means or Gaussian mixture models. However, these traditional approaches often fail to capture the complex, nonlinear relationships inherent in multi-scale genomic data.

## 2.3 Contrastive learning

Contrastive learning has recently emerged as a powerful paradigm for learning discriminative representations from unlabeled data. By maximizing agreement between different views or augmentations of the same sample, contrastive methods can uncover meaningful latent structure in the data without requiring class labels. Chen *et al*. (2023) applied contrastive learning for cancer subtyping across 10 multi-omic datasets. Wang *et al*. (2025) introduced CLCluster, a redundancy-reduction contrastive learning model integrating copy number variation, DNA methylation, gene expression, miRNA expression, and alternative splicing to cluster 33 cancer types. Zhao *et al*. (2023) proposed a contrastive learning-based framework for cancer subtyping using multi-omics data. However, all these efforts were limited to the patient level.

Although contrastive learning has shown promise in biomedical imaging and patient-level omics, it has not yet been applied to cohort-level cancer clustering using mutation data. Our work fills this gap by introducing MS-ConTab, a fully unsupervised, multi-scale contrastive learning framework that employs TabNet encoders and the NT-Xent loss to integrate gene-level and chromosome-level mutation signatures. These learned embeddings enable robust clustering of cancer types and reveal novel insights into pan-cancer mutational processes.

# 3 Methods

We propose a fully unsupervised, contrastive learning framework for cohort-level cancer clustering based on protein-coding single-nucleotide substitution patterns. The goal is to learn meaningful representations from somatic coding mutations at both the gene and chromosome levels and use these embeddings to uncover biologically coherent cancer groupings. We define gene level as mutations within coding regions of genes (excluding introns). To understand whether each cancer is associated with mutations in certain genomic regions, we performed what we define a “chromosome level” analysis. In this analysis, we associated mutation patterns in each chromosome with cancer type (**Supplementary Figure 1A, B**). For both analyses, we focused on the 25 most frequently mutated genes in each cancer. As illustrated in **Figure 1**, our framework consists of three key stages: (1) Aggregation of gene-level and chromosome-level mutation features from curated COSMIC somatic variant data, focusing on the top 25 most frequently mutated protein-coding genes per cancer type and their nucleotide substitution frequencies. (2) Dual TabNet encoders trained using the NT-Xent contrastive loss to align gene-level and chromosome-level embeddings into a unified latent space, and (3) Application of clustering algorithms on the learned embeddings to identify distinct cancer type groupings. This multi-scale design enables the model to capture both localized gene-specific and genome-wide chromosomal mutational signatures, facilitating robust, biologically interpretable cohort-level clustering without supervision. The details of the frameworks are described in the next sections.

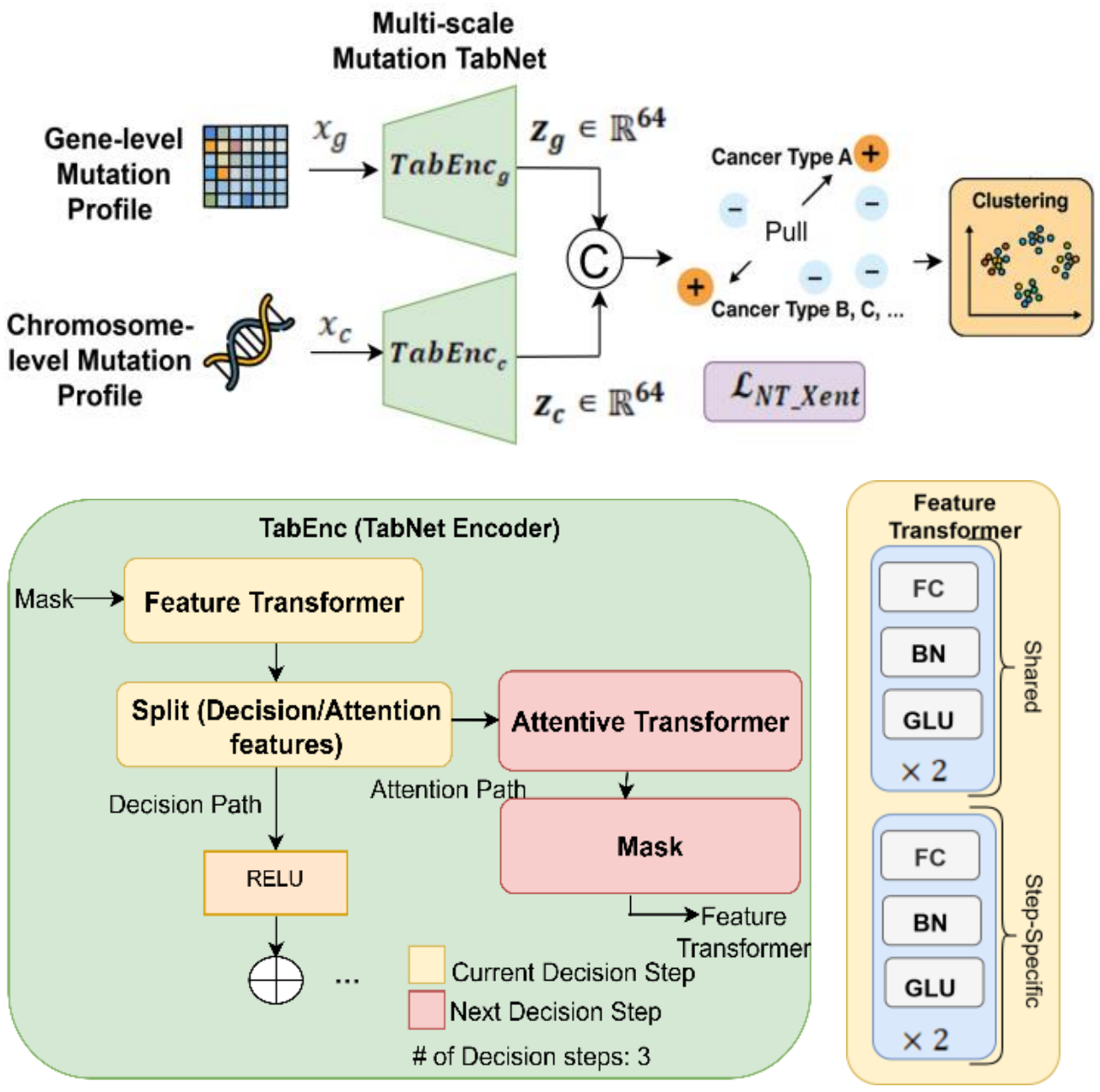

**Figure 1. Overall proposed framework (MS-ConTAB)**

## 3.1 Data Acquisition and Preprocessing

Mutation data for 43 distinct cancer types were curated from the COSMIC database (Tate *et al*. 2019). We retained only single-nucleotide substitutions, removed all alternative transcript data to prevent over-counting (e.g. all gene identifiers whose gene names end in “_ENST”) and excluded non–protein-coding variants to focus on high-confidence, protein-coding mutations (**Supplementary Figure 1A, B**). For each cancer type, two complementary “views” of mutation patterns were constructed: 1) a gene-level signature and a 2) chromosome-level signature. The gene-level feature vector was generated by identifying the top 25 most frequently mutated protein-coding genes within each cancer type and quantifying the counts of 12 specific nucleotide substitutions (A>T, C>G, etc.) for each gene (**Supplementary Figure 1C**). This yielded a 25 × 12 matrix per cancer type, which was then flattened to a 300-dimensional vector. For the chromosome-level modality, we computed the counts of all 12 nucleotide substitution types for each chromosome, resulting in a vector with dimensionality equal to the number of chromosomes. To account for differences in chromosome size, these counts were normalized by chromosome length.

## 3.2 TabNet Encoder

In our framework, TabNet (Arik and Pfister 2021) is employed as the encoder to transform raw mutation features into low-dimensional, informative cancer-type representations. TabNet is a deep learning architecture specifically designed for tabular data, leveraging a sequential attention mechanism to dynamically select the most relevant features at each decision process. This makes it especially well-suited for high-dimensional and heterogeneous genomic inputs, where some features (e.g., specific gene-level nucleotide changes) are highly informative while others are redundant or noisy.

Each cancer type is represented by two complementary input views: 1) gene-level mutation vector: 300 features derived from the top 25 most frequently mutated protein-coding genes (25 genes × 12 nucleotide substitutions), and 2) chromosome-level mutation vector: 288 features representing aggregated nucleotide substitutions across 24 chromosomes (24 chromosomes × 12 substitution types) (Supplementary Figure 1). Separate TabNet encoders are applied to each view using an identical architecture composed of multiple decision steps. In each step, TabNet generates a sparse attention mask (via sparsemax activation), selecting a subset of features to transform through feature transformer blocks, and progressively aggregates these transformations across steps to produce a final latent representation. For each view, TabNet produces a 64-dimensional latent vector, which is then mapped to a 64-dimensional contrastive embedding space through a shallow projection head. This compact dimensionality was selected to enforce well-regularized embeddings and reduce overfitting, making the model more suitable for the relatively small cohort-level dataset. This design enables the model to learn view-specific, interpretable representations, allowing different cancer types to attend to different sets of genes or chromosomes. The gene-level and chromosome-level TabNet encoders are trained jointly within a contrastive learning framework, encouraging alignment of embeddings for the same cancer type while enforcing separation from embeddings of other cancer types.

## 3.3 Contrastive Loss Function (NT-Xent)

To train the TabNet encoders in a self-supervised fashion, we adopt a contrastive learning strategy that encourages alignment between the two complementary views of each cancer type: gene-level and chromosome-level mutation signatures. Specifically, we employ the Normalized Temperature-scaled Cross-Entropy Loss (NT-Xent), originally introduced in SimCLR for image representation learning, but here adapted for multi-view genomic data. While both our framework and the SimCLR baseline use NT-Xent, MS-ConTab fundamentally differs in architecture and problem framing. Unlike SimCLR, which applies a single-view MLP encoder, MS-ConTab leverages dual TabNet encoders to integrate gene- and chromosome-level mutation signatures, enabling multi-scale, interpretable representation learning tailored for cohort-level cancer clustering.

Each cancer type is represented by two input modalities: a gene-level vector and a chromosome-level vector. These views are passed through their respective TabNet encoders, resulting in two latent embeddings, $z_g \in \mathbb{R}^{64}$ and $z_c \in \mathbb{R}^{64}$, for the gene and chromosome modalities, respectively. The objective of contrastive learning is to bring these two embeddings of the same cancer type (positive pair) closer in the embedding space, while pushing them away from embeddings of other cancer types in the same batch (negative pairs).

The NT-Xent loss for an anchor embedding $z_i$ and its corresponding positive pair $z_j$ is defined as:

$$\mathcal{L}_i = -\log \frac{exp\left(\frac{S(z_i, z_j)}{\tau}\right)}{\sum_{k=1}^{2N} 1_{[k \neq j]} exp\left(\frac{S(z_i, z_k)}{\tau}\right)} \qquad (1)$$

where $S(z_i, z_j) = \frac{z_i \cdot z_j}{\|z_i\|\|z_j\|}$ is the cosine similarity between normalized embeddings, $\tau$ is the temperature hyperparameter (set to 0.5 in our experiments), and $2N$ is the total number of views in a batch (i.e., $N$ cancer types with two embeddings each). The total loss is averaged across all positive pairs in a batch. This formulation promotes clustering of corresponding cancer-type representations from the two views, while enforcing separation from unrelated types. The total batch loss is averaged across all views:

$$\mathcal{L}_{NT_Xent} = \frac{1}{2N} \sum_{i=1}^{2N} \mathcal{L}_i \qquad (2)$$

This formulation enforces high similarity between gene-level and chromosome-level embeddings of the same cancer type, while maintaining separation between embeddings of different types. As a result, the model learns modality-invariant, discriminative representations that support robust clustering in downstream analyses.

# 4 Experiments and Results

## 4.1 Experimental Setup

All models were implemented in Python 3.9 and trained on a workstation with an Intel CPU (32 GB RAM). Deep learning components were implemented in PyTorch and downstream analyses, including clustering, dimensionality reduction, and visualization were performed using scikit-learn, umap-learn, and matplotlib. Our proposed MS-ConTab model uses dual TabNet encoders, each producing a 64-dimensional latent representation followed by a 64-dimensional projection head for contrastive embedding. The latent dimension was set to 64 to balance representational capacity with the small sample size, mitigating overfitting risks.

Models were trained for 100 epochs using the Adam optimizer (learning rate = 1e-3, batch size = 8) with the NT-Xent contrastive loss (temperature = 0.5). All experiments used a fixed random seed (42) for reproducibility. For comparison, we evaluated five baselines: NMF, hierarchical clustering, a shallow autoencoder, SimCLR with a single-view MLP encoder, and DeepCluster. These baselines provide a spectrum of classical, deep learning, and contrastive approaches for unsupervised cancer-type clustering.

## 4.2 Two well-separated clusters of cancers

The projection of the learned cancer embeddings revealed two well separated clusters among the 42 cancer types. Specifically, we visualized the embeddings using both t-SNE (**Figure 2A**) and UMAP (**Figure 2B**), both of which demonstrated a clear two-cluster structure. To further validate this observation, we applied k-means clustering (**Figure 2C**) on the learned embedding space, which robustly recapitulated the same two-cluster pattern. We listed the names of cancer types in two clusters (**Figure 2D**). This consistency across visualization and clustering methodsindicates that the separation is a stable property of the learned representations rather than an artifact of a single dimensionality reduction technique.

Cluster 1 predominantly comprises solid epithelial and parenchymal tumors originating in well-defined organs such as the large intestine, lung, liver, kidney, pancreas, stomach, and thyroid, as well as soft-tissue sarcomas and central nervous system malignancies. In contrast, Cluster 2 groups together endocrine, reproductive, and hematopoietic cancers. This includes hormone-sensitive tissues (breast, fallopian tube, female genital tract, prostate—though partially distinct), true endocrine organs (parathyroid, pituitary; thyroid remains in Cluster 1), hematolymphoid neoplasms, and reproductive tract subtypes (vulva, vagina, testis, placenta, uterine adnexa) (**Figure 2D**). These cancers may be influenced by lineage-specific developmental or hormonal programs in addition to mutation processes.

Further analysis of within- and between-cluster similarities revealed high intra-cluster coherence (mean cosine similarity: 0.591 for Cluster 1; 0.965 for Cluster 2) and a negative mean similarity between clusters (-0.287), confirming that the two clusters are both internally consistent and well separated in the embedding space. The most central cancer types, identified as cluster prototypes by cosine similarity, were Thyroid for Cluster 1 and Haematopoietic and Lymphoid for Cluster 2.

To complement the clustering results, we performed a nearest-neighbor analysis in the learned embedding space, identifying the three most similar

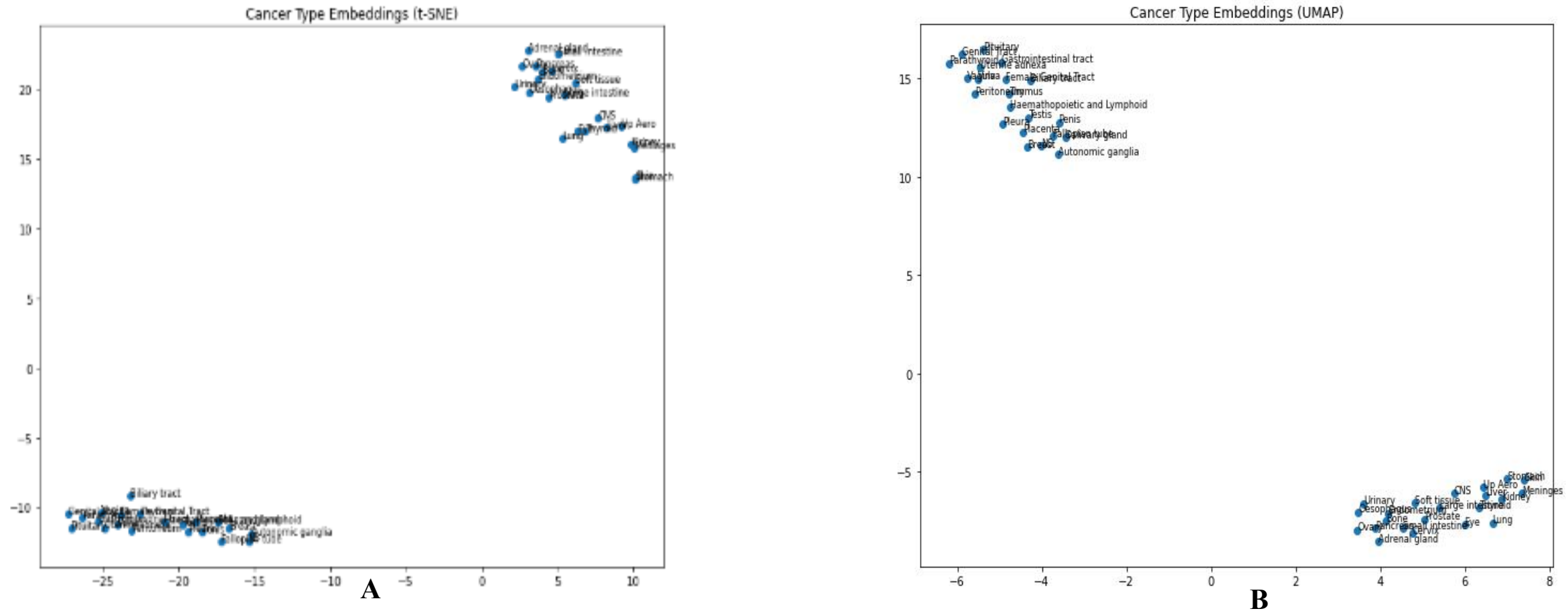

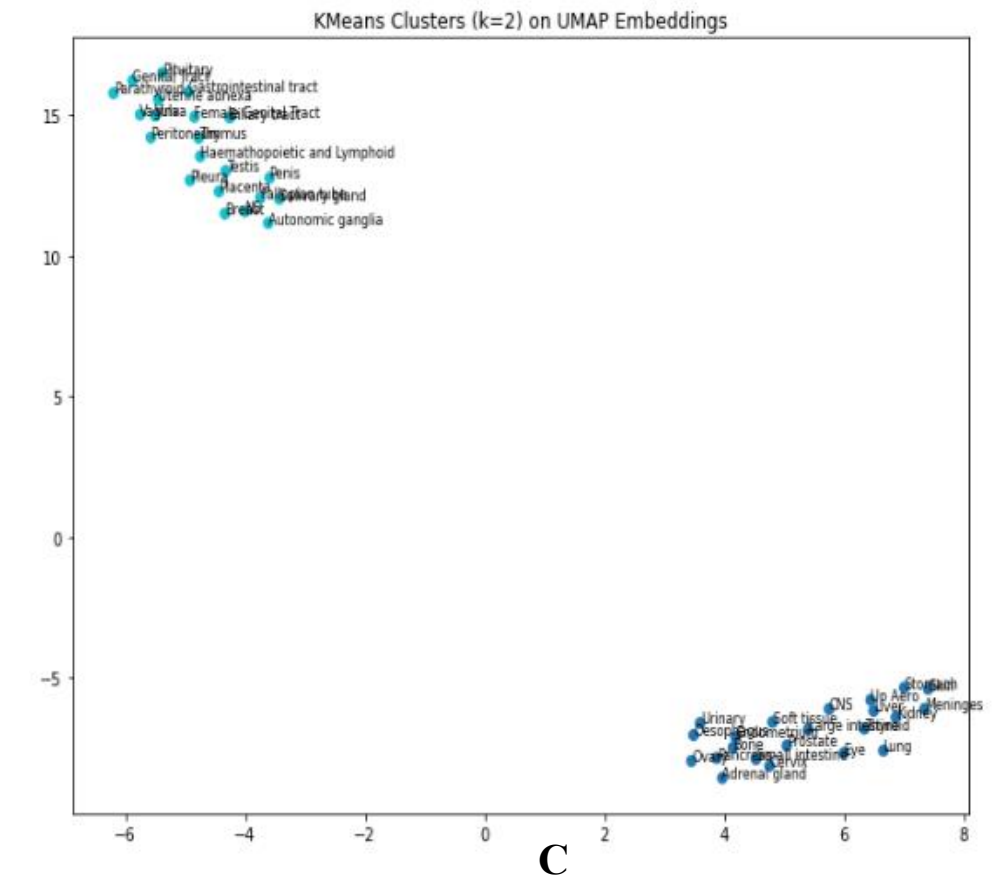

| Cluster 1 | Cluster 2 |
|---|---|
| Adrenal gland | Autonomic ganglia |
| Bone | Biliary tract |
| CNS | Breast |
| Cervix | Fallopian tube |
| Endometrium | Female Genital Tract |
| Eye | Gastrointestinal tract |
| Kidney | Genital Tract |
| Large intestine | Haematopoietic and Lymphoid |
| Liver | NS |
| Lung | Parathyroid |
| Meninges | Penis |
| Oesophagus | Peritoneum |
| Ovary | Pituitary |
| Pancreas | Placenta |
| Prostate | Pleura |
| Skin | Salivary gland |
| Small intestine | Testis |
| Soft tissue | Thymus |
| Stomach | Uterine adnexa |
| Thyroid | Vagina |
| Up Aero | Vulva |
| Urinary | |

D

**Figure 2. Visualization and clustering of cancer type embeddings**. Two-dimensional projections of the learned cancer embeddings using **A**. t-SNE, **B**. UMAP, and **C**. UMAP with KMeans cluster assignments (clusters are shown in two diffetent colors) (k=2). **D** lists the names of cancer types grouped into two clusters based on MS-ConTab embeddings. Each point represents a cancer type, labeled accordingly. The projections consistently

cancer types for each entity (**Supplementary Table 1**). The results reveal biologically coherent relationships: for instance, ovary, endometrium, and pancreas cancers frequently appear as mutual neighbors, reflecting shared developmental and mutational programs, while hematopoietic and lymphoid malignancies cluster near thymus and pleura, consistent with their immune-related biology. Interestingly, several cross-lineage proximities also emerged (e.g., breast clustering with salivary gland and pleura), suggesting shared underlying mutational processes not captured by tissue-of-origin alone. This analysis provides fine-grained insight into inter-cancer relationships, complementing the broader two-cluster structure observed in our main results.

even more distinct when projected into two dimensions. Similarly, the Davies-Bouldin (DB) Index is 0.655 and 0.136 (lower is better) in original and UMAP prpjection , respectively, reflecting improved intra-cluster compactness and inter-cluster separation. The Calinski-Harabasz (CH) Index rose sharply from 43.1 to 1862.0, suggesting a dramatic increase in between-cluster dispersion relative to within-cluster variance. Collectively, these metrics confirm that MS-ConTab generates coherent and biologically meaningful clusters, with UMAP further enhancing their separability for visualization and interpretability.

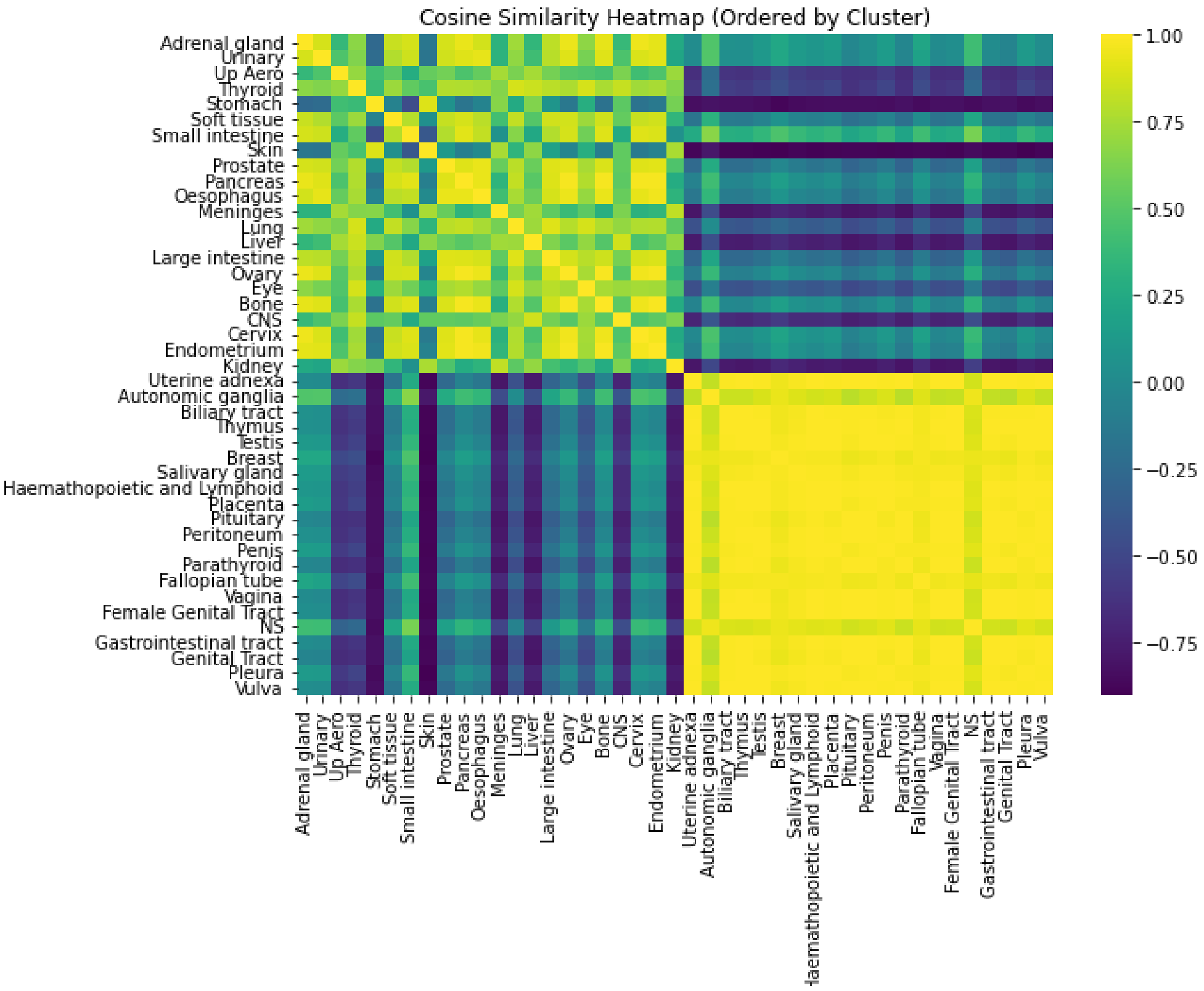

**Figure 3. Cosine similarity heatmap of cancer type embeddings**. This heatmap displays the pairwise cosine similarity between the learned embeddings of all 43 cancer types, with rows and columns ordered to group cancers by their KMeans cluster assignment. Yellow regions indicate high similarity (close to 1), while blue or purple regions indicate low or negative similarity. Within-cluster blocks (upper-left and lower-right quadrants) show strong internal coherence, with most pairwise similarities being high, especially in Cluster 2, which is highly homogeneous. Between-cluster blocks (off-diagonal) display much lower, and often negative, similarity values, reflecting clear separation between the two major cancer groups. The sharp contrast between the within-cluster and between-cluster regions visually confirms that the contrastive learning approach produced highly distinct cancer groupings, in line with quantitative metrics such as the silhouette score and mean within-/between-cluster similarity.

The cosine similarity heatmap of the learned cancer-type embeddings is shown in **Figure 3**. Rows and columns are ordered by cluster assignment, with each cell representing the pairwise cosine similarity between prominent diagonal blocks emerge: cancers within Cluster 1 (predominantly solid epithelial tumors) show uniformly high intra-cluster similarity, while cancers within Cluster 2 (hormone-related and hematopoietic malignancies) exhibit a similarly coherent pattern. In contrast, the off-diagonal regions are largely low or negative, indicating that cancers from different clusters share little mutational embedding structure. These patterns confirm both the strong internal coherence of each cluster and their clear separation in the learned embedding space.

### 4.3 Quantitative Evaluation Confirms Cluster Robustness

To assess clustering performance, we assessed the quality of groupings in the original MS-ConTab embedding space and their 2-D UMAP projection (**Table 1**). The Silhouette Score is 0.561 in the original space and 0.906 in UMAP projection, indicating that while MS-ConTab embeddings already formed reasonably well-defined clusters, their separation became

**Table 1.** Clustering quality metrics for MS-ConTab embeddings. Higher Silhouette and Calinski–Harabasz (CH) indices and lower Davies–Bouldin (DB) indices indicate better clustering performance.

| Metric | UMAP | Original Embeddings |
|---|---|---|
| **Silhouette Score (k=2)** | 0.906 | 0.561 |
| **DB Index** | 0.136 | 0.655 |
| **CH Index** | 1862.0 | 43.1 |

### 4.4 Mutation Spectrum and Chromosomal Load Differences Between Clusters

We assessed cluster-level differences in mutational processes by examining gene level as well as chromosome level nucleotide substitutions. **Figure 4A** illustrates the average mutation counts for each of the 12 possible base–substitution types, for each cluster (Supplementary Figure 1C). Both clusters are dominated by C→T and G→A transitions, a pattern characteristic of spontaneous cytosine deamination and related endogenous processes. This is not entirely unexpected because transitions are are more likely to occur than transversions (Vogel and Kopun, 1977**).** However, these substitutions occur at markedly higher amplitudes in Cluster 1, reflecting a molecular difference between these two clusters. Cluster 2 demonstrates the same substitution preferences but with substantially lower overall counts, suggesting a lower mutation load.Interestingly, G→T transversions are slightly more frequent in Cluster 2 compared to Cluster 1, hinting at possible cluster-specific mutational mechanisms. Importantly, the observation that complementary substitutiosn are not equal (e.g. G>A = C>T) suggests that at least some mutation processes are replication independent (Abascal F *et al. 2021* , Helleday T *et al.* 2014).

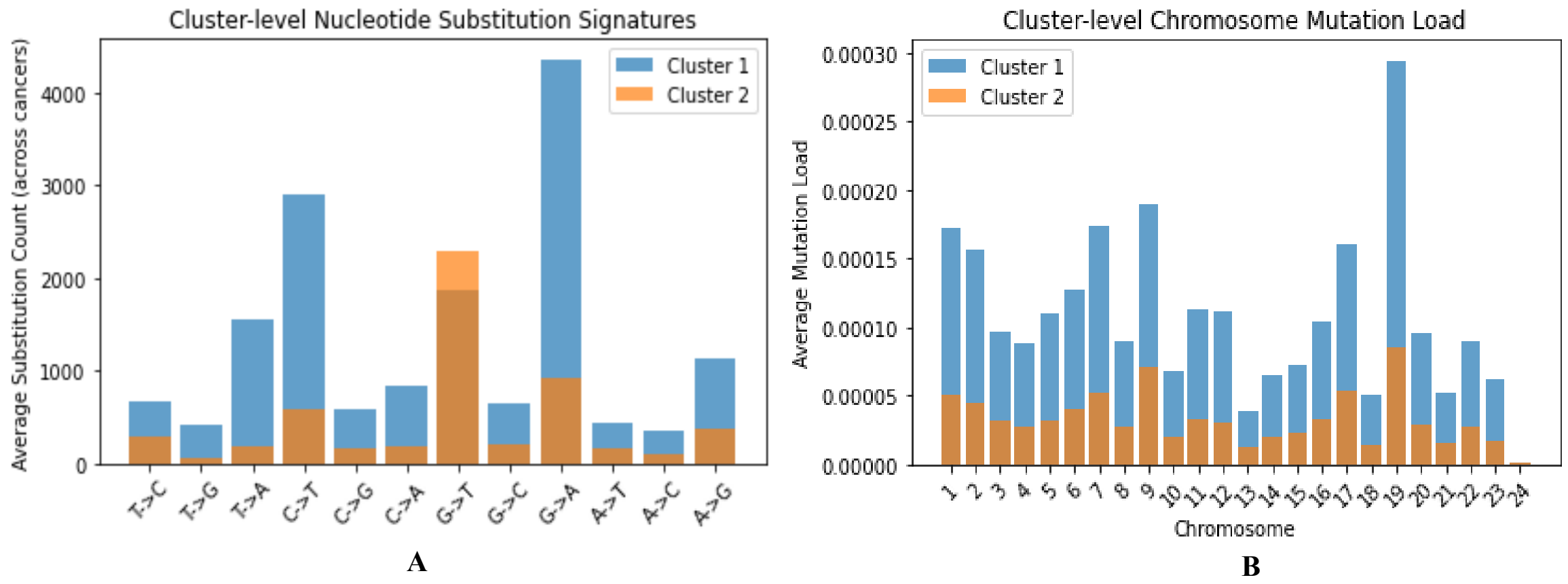

**Figure 4. Comparison of mutation patterns between cancer clusters (overlaid bars)**. **A**) Cluster-level nucleotide substitution signatures. Bars show the average count of each substitution type across cancers in cluster 1 (blue) and cluster 2 (orange). Cluster 1 exhibits higher rates of G>A and C>T substitutions, while cluster 2 displays a higher rate in G>T. **B**) Cluster-level chromosome mutation load. Bars indicate the average mutation load per chromosome for each cluster, with cluster 1 showing consistently higher mutation burden across all chromosomes compared to cluster 2.

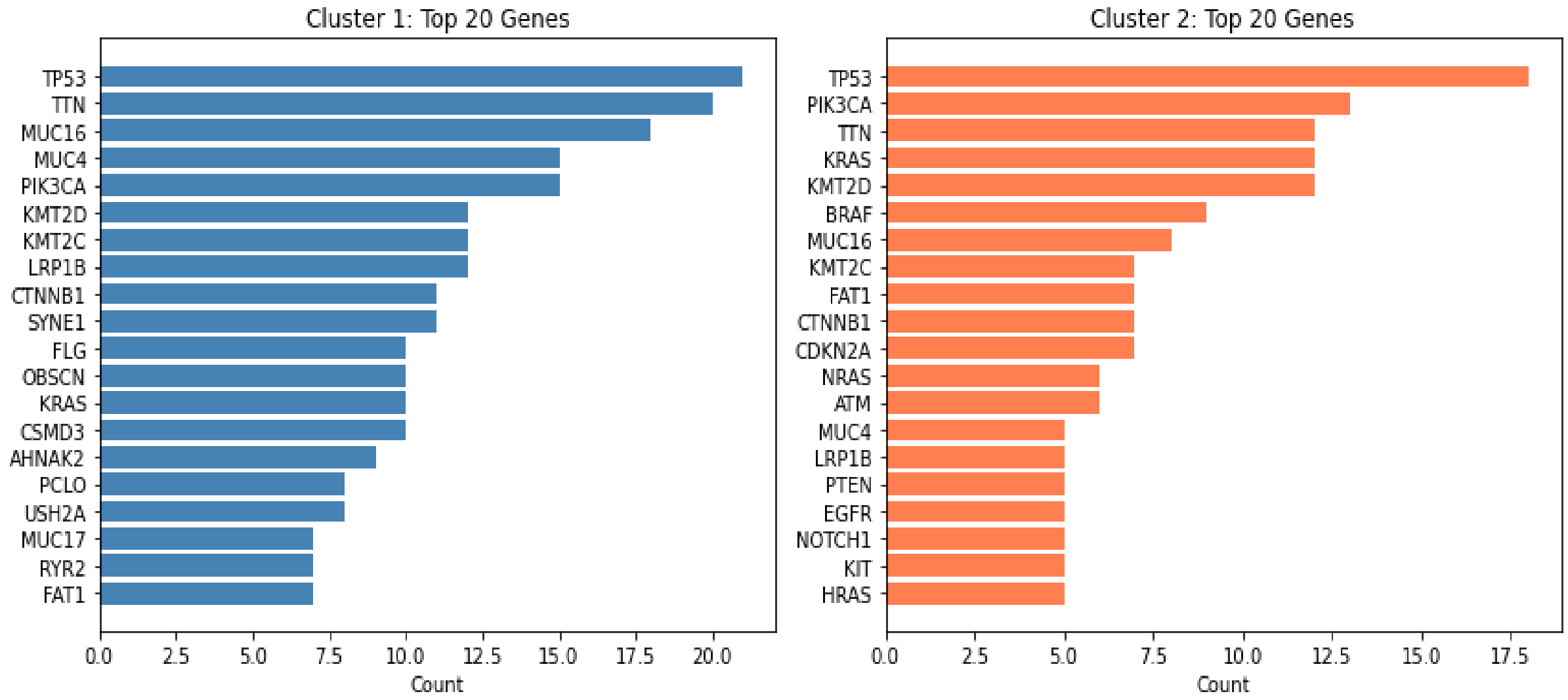

**Figure 5. Top recurrently mutated genes distinguishing cancer clusters.** Left plot: The top 20 most frequently mutated genes across cancer types in Cluster 1, with bar length indicating the number of cancers within the cluster in which each gene is among the most mutated. Right plot: The corresponding top mutated genes for Cluster 2. Distinct sets of genes characterize each cluster, highlighting differences in mutational drivers and biological pathways between groups of cancers.

**Figure 4B** shows the average per-chromosome mutation load for each cluster, normalized by chromosome length. Cluster 1 exhibits a markedly higher mutation burden across nearly all chromosomes compared to Cluster 2, consistent with the elevated genomic instability characteristic of solid tumors. In particular, chromosomes 19, 7, 9, and 1 stand out in Cluster 1, with chromosome 19 reaching the highest mean load (~$3.0\times10^{-4}$), suggesting that mutational processes operate unequally in the genome for cluster 1 cancers. By contrast, Cluster 2 shows a uniformly low and relatively flat mutation load (~$0.5–1.0\times10^{-4}$) across chromosomes, indicating a more stable and potentially homogeneous mutation landscape. These differences reflect variation in mutational processes across clusters. a more stable and potentially homogeneous mutation landscape. These differences reflect variation in mutational processes across clusters.

## 4.5 Top Mutated Genes Distinguish Cancer Clusters

Analysis of the most recurrently mutated genes revealed striking differences between the two cancer clusters (**Figure 5**). Not unexpectedly, TP53 dominates both clusters, underscoring its near-ubiquitous role as a pan-cancer driver. PIK3CA emerges as a key driver in Cluster 2, highlighting the centrality of the PI3K pathway in hormone-related and hematopoietic cancers. TTN and MUC16 follow closely in Cluster 1, but these genes have been previously reported to be potential false positives (Lawrence MS *et al.* 2013**)** and our interpretation of their role here is also cautious.

Cluster 1 is enriched for structural and adhesion-related genes such as FLG, OBSCN, AHNAK2, and USH2A, consistent with epithelial and connective-tissue malignancies. It also features classic carcinoma drivers (KRAS, KMT2D, KMT2C, LRP1B, CTNNB1). Cluster 2, by contrast, features unique regulators of cell-cycle control and signaling, including CDKN2A, NRAS, ATM, PTEN, EGFR, NOTCH1, KIT, and HRAS, reflecting the diverse driver pathways of reproductive and hematopoietic malignancies. Overall, these paired barplots reveal a shared backbone of universal drivers (TP53, TTN, MUC16, KRAS) alongside cluster-specific genes that shape the distinct mutational landscapes of solid versus hormone-/immune-related cancer types.

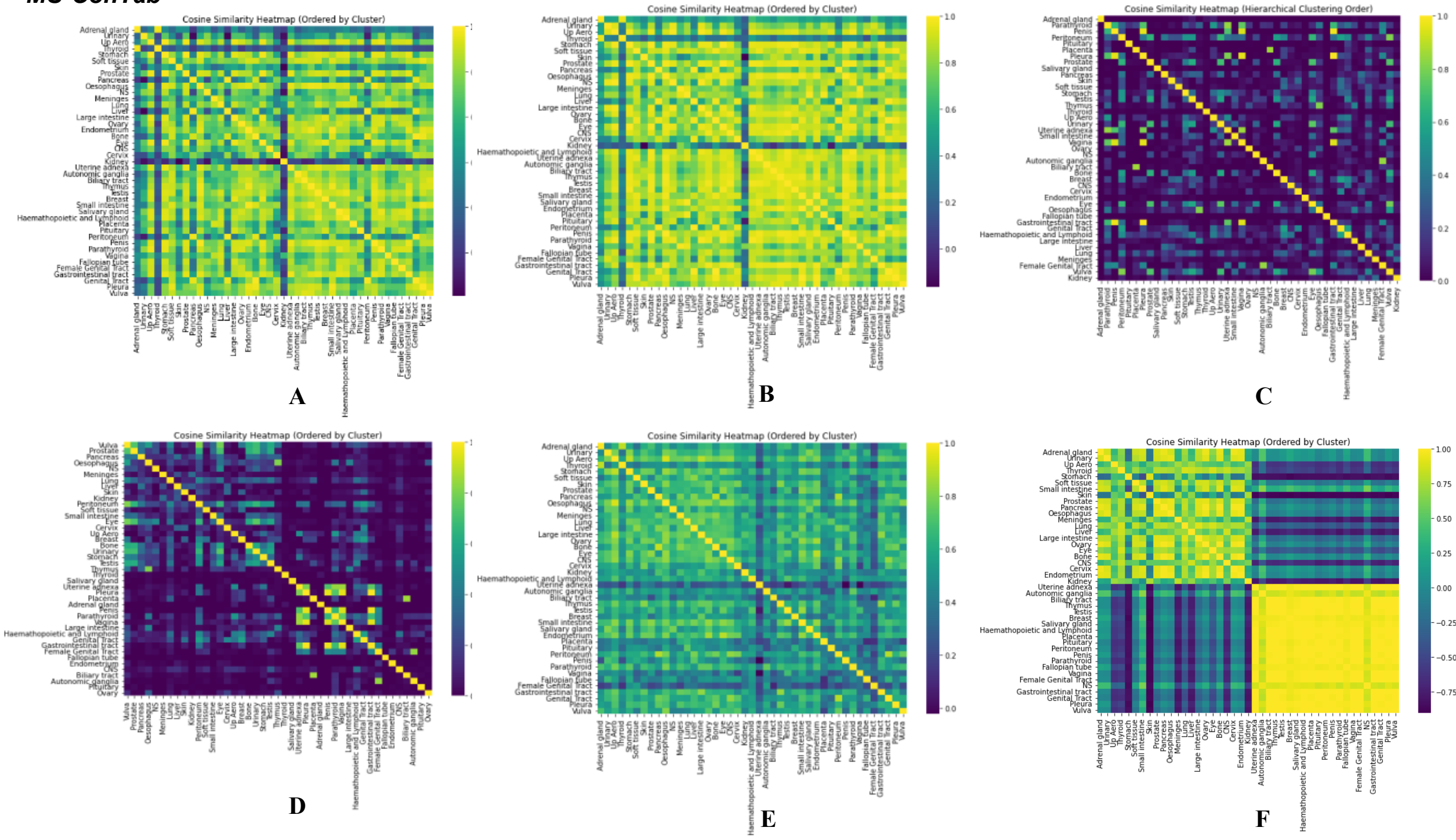

**Figure 7. Comparison of Cosine similarity heatmaps of cancer-type embeddings across State-of-the-art methods:** A) Autoencoder, B) deepCluster, C) Hierarchical, D) NMF, E) SimCLR F) MS-ConTab. Each heatmap shows the pairwise cosine similarities between cancer types, reordered by cluster assignments where applicable. Brighter colors indicate higher similarity, and darker colors indicate lower similarity.

To further characterize the clusters, we compared the complete sets of mutated genes across them (**Supplementary Table 2**). We identified 117 genes shared between clusters, including well-known pan-cancer drivers such as TP53, KRAS, PIK3CA, PTEN, and ARID1A. Cluster 1 exhibited 73 unique genes (e.g., FOXL2, STK11, KEAP1), many linked to epithelial and structural functions. Cluster 2 contained 148 unique genes (e.g., JAK2, RUNX1, MYC, EZH2), reflecting enrichment for transcriptional regulation, hematopoietic signaling, and developmental pathways. This overlap and divergence highlight a shared backbone of pan-cancer drivers alongside distinct, cluster-specific gene programs. **Figure 6** shows a Venn diagram of recurrently mutated genes across the two clusters.

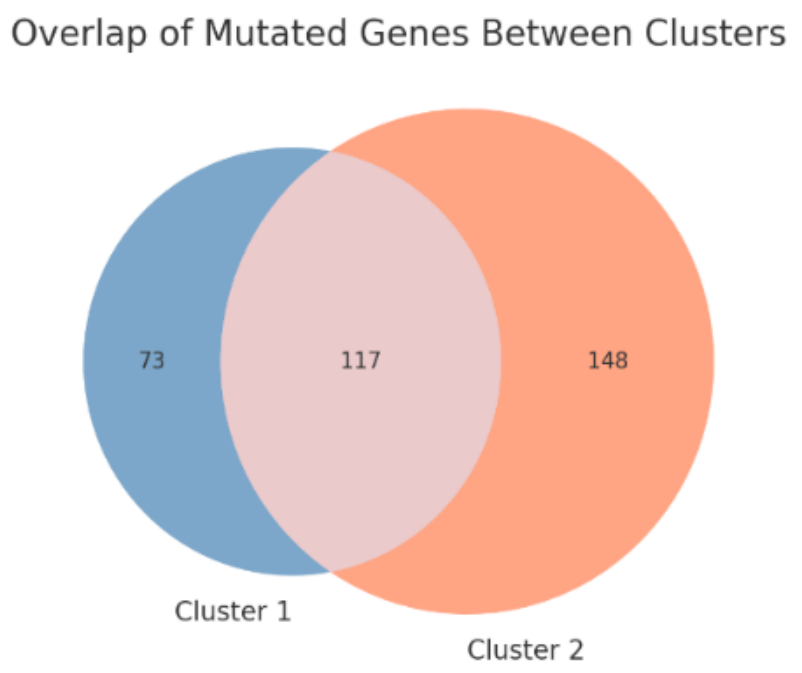

**Figure 6. Venn diagram showing the overlap of recurrently mutated genes between the two cancer clusters.** The diagram illustrates 117 genes shared by both clusters, alongside 73 unique to Cluster 1 (solid-tumor–enriched) and 148 unique to Cluster 2 (hormone-related and hematopoietic malignancies), highlighting both common pan-cancer drivers and cluster-specific mutational signatures.

### 4.6 Comparison to Baselines

We compared the performance of our proposed MS-ConTab against several baseline approaches to assess the quality and discriminative power of the learned cancer-type embeddings. Specifically, we assessed embeddings generated using NMF, Autoencoders, DeepCluster (Caron *et al*. 2018), SimCLR-style MLP-based contrastive model (Chen *et al*. 2020), and hierarchical clustering applied directly to the concatenated gene-level and chromosome-level mutation features, all using the same downstream clustering pipeline. **Table 2** summarizes the clustering performance of six embedding/clustering methods using k-means (k=2) in their original feature space with corressponding results after UMAP projection provided in **Supplementary Table 3.** The proposed TabNet + contrastive learning approach achieved a strong balance across all metrics, with a Silhouette score of 0.561, a low DB index (0.655), and the highest CH index (43.1), indicating MS-ConTab capture more compact and well-separated cluster structure than competing methods.

**Table 2.** Comparison of clustering quality across embedding methods using k-means (k=2) in their original learned space. Higher Silhouette and Calinski–Harabasz (CH) indices and lower Davies–Bouldin (DB) indices indicate better clustering quality.

| Embedding Method | Silhouette Score (↑) | DB Index (↓) | CH Index (↑) |
|---|---|---|---|
| MS-ConTab (ours) | 0.561 | 0.655 | 43.106 |
| NMF (k=43) | 0.141 | 3.011 | 2.680 |
| Hierarchical (Ward) | 0.697 | 0.194 | 16.275 |
| Autoencoder (AE) | 0.187 | 1.866 | 2.385 |
| SimCLR (MLP-based)(Chen *et al*. 2020) | 0.239 | 1.378 | 4.721 |
| DeepCluster (Caron *et al*. 2018) | 0.216 | 1.60 | 3.16 |

While hierarchical clustering achieved the highest Silhouette score (0.697), its clusters were highly imbalanced, as reflected by its substantially lower CH index (16.3). Baseline methods such as NMF, Autoencoder, SimCLR, and DeepCluster exhibited weaker cluster quality, with notably lower Silhouette and CH scores and higher DB indices. These results demonstrate that MS-ConTab produces more discriminative and biologically meaningful embeddings compared to both traditional and deep learning baselines.

The heatmaps in **Figure 7** reveal substantial differences in the clustering structures produced by different embedding strategies. Autoencoder and DeepCluster embeddings (A, B) produce relatively diffuse similarity structures, suggesting weaker separation between cancer types. Hierarchical clustering (C) shows a more distinct block pattern, reflecting its tendency to isolate a few highly similar groups, though sometimes at the expense of overall balance. NMF (D) exhibits partially structured clusters but with less pronounced separation compared to advanced methods. SimCLR (E) improves separation compared to simpler baselines, but its clusters are less cohesive than those of TabNet. Although the SimCLR also leverages NT-Xent loss, its MLP-based design achieved significantly lower clustering quality. In contrast, MS-ConTab (F) generates the clearest and most well-separated blocks, highlighting its ability to produce biologically meaningful and discriminative embeddings. Collectively, these results reinforce that TabNet embedding with contrastive learning yields the most interpretable and biologically coherent representations of cancer-type mutation profiles.

## 5 Conclusion

In this study, we introduced MS-ConTab, a novel multi-scale contrastive learning framework for cohort-level cancer clustering based on somatic nucleotide substitution profiles. By leveraging complementary gene-level and chromosome-level mutation views and training dual TabNet encoders with NT-Xent contrastive loss, our method learns robust cancer-type embeddings without supervision. Comprehensive evaluation against classical and deep learning baselines demonstrated that MS-ConTab produces biologically meaningful clusters that align with known tissue lineages and mutational processes. In particular, it effectively distinguishes solid epithelial tumors from hormone-related and hematopoietic malignancies, revealing consistent intra-cluster structure and distinct genomic signatures across substitution types, chromosomal loads, and top mutated genes.

Our findings highlight the power of contrastive representation learning for uncovering pan-cancer relationships beyond individual-level subtyping, especially when applied to interpretable, multi-scale mutation data. MS-ConTab sets the foundation for future applications in pan-cancer taxonomy, drug repurposing, and mutation signature discovery. Future work could extend this approach to include additional genomic modalities (e.g., structural variants, non-coding annotations) and test its utility in other biological domains.

## References

Abascal F, Harvey LMR, Mitchell E *et al*. Somantic mutation landscapes at single-molecule resolution.*Nature* 2021;**593**:405-410.

Al-Azani S, Alkhnbashi OS, Ramadan E *et al*. Gene expression-based cancer classification for handling the class imbalance problem and curse of dimensionality. *Int J Mol Sci* 2024;**25**:2102.

Alexandrov LB, Kim J, Haradhvala, NJ *et al.* The repertoire of mutational signatures in human cancer. *Nature* 2020;**578**:94–101.

Alexandrov LB, Nik-Zainal S, Wedge DC *et al*. Signatures of mutational processes in human cancer. Nature 2013;**500**:415-421.

Arik SÖ, Pfister T. TabNet: Attentive Interpretable Tabular Learning. In: *AAAI Conf Artif Intell* 2021;**35**:6679-6687.

Arora A, Olshen AB, Seshan VE et al. Pan-cancer identification of clinically relevant genomic subtypes using outcome-weighted integrative clustering. *Genome Medicine* 2020;**12**:110.

Bailey MH, Tokheim C, Porta-Pardo E *et al*. Comprehensive characterization of cancer driver genes and mutations. *Cell* 2018;**173**:371-385.e18.

Brunet JP, Tamayo P, Golub TR *et al*. Metagenes and molecular pattern discovery using matrix factorization. *Proc Natl Acad Sci USA* 2004;**101**:4164-4169.

Caron M, Bojanowski P, Joulin A *et al*. Deep clustering for unsupervised learning of visual features. In: *Proceedings of the European Conference on Comput Vission (ECCV)*, pp.132-149, 2018.

Chen T, Kornblith S, Norouzi M *et al*. A simple framework for contrastive learning of visual representations. In: Proceddings of the 37th International Conference on Machine Learning, pp.1597-1607, 2020.

Chen W, Wang H, Liang C. Deep multi-view contrastive learning for cancer subtype identification. *Brief Bioinformatics* 2023;**24**:bbad282.

Dou Y, Mirzaei G. MO-GCAN: Multi-Omics Integration based on Graph Convolutional and Attention Networks. *Bioinformatics* 2025;**41**:btaf405.

Helleday T, Eshtad S, Nik-Zainal S. Mechanisms underlying mutational signatures in human cancers. *Nature* 2014; **15**: 585-598.

Hoadley KA, Yau C, Wolf DM *et al*. Multiplatform analysis of 12 cancer types reveals molecular classification within and across tissues of origin. *Cell* 2014;**158**:929-944.

Lawrence MS, Stojanov Petar, Polak P *et al*. Mutational heterogeneity in cancer and the search for new cancer-associated genes. *Nature* 2013; **499**: 214-218.

Mirzaei G, Petreaca RC. Distribution of copy number variations and rearrangement endpoints in human cancers with a review of literature. *Mutat Res* 2022;**824**:111773.

Mirzaei G. Constructing gene similarity networks using co-occurrence probabilities. *BMC Genomics* 2023;**24**:697.

Mirzaei G. GraphChrom: A novel graph-based framework for cancer classification using chromosomal rearrangement endpoints. *Cancers* 2022;**14**:3060.

Mo Q, Wang S, Seshan VE *et al*. Pattern discovery and cancer gene identification in integrated cancer genomic data. *Proc Natl Acad Sci USA* 2013;**110**:4245-4250.

Orrantia-Borunda E, Anchondo-Nuñez P, Acuña-Aguilar LE, *et al*. Subtypes of Breast Cancer. In: Mayrovitz HN, editor. *Breast Cancer* [Internet]. Brisbane (AU): Exon Publications; 2022 Aug 6. Chapter 3.

Raj A, Mirzaei G. Multi-armed bandit approach for multi-omics integration. In: *2022 IEEE International Conference on Bioinformatics and Biomedicine (BIBM)*, *IEEE*, Las Vegas, NV, USA, pp. 3130-3136, 2022.

Raj A, Petreaca RC, Mirzaei G. Multi-Omics Integration for Liver Cancer Using Regression Analysis. *Curr Issues Mol Biol* 2024;**46**:3551-3562.

Shen R, Olshen AB, Ladanyi M. Integrative clustering of multiple genomic data types using a joint latent variable model with application to breast and lung cancer subtype analysis. *Bioinformatics* 2009;**25**:2906-2912.

Sidaway P. Glioblastoma subtypes revisited. *Nature Review Clinical Oncology* 2017;**14**:587.

Tate JG, Bamford S, Jubb HC *et al*. COSMIC: the Catalogue Of Somatic Mutations In Cancer. *Nucleic Acids Res* 2019;**47**:D941-D947.

The Cancer Genome Atlas Research Network, Weinstein JN, Collisson EA, *et al*. The Cancer Genome Atlas Pan-Cancer analysis project. *Nature Genetics* 2013;**45**:1113-1120.

Vogel F, Kopun M. Higher Frequencies of Transitions among Point Mutations. *Journal of Molecular Evolution* 1977; **9**:159-180.

Wang H, Zhang Y, Li W *et al*. CLCluster: A redundancy-reduction contrastive learning-based clustering method of cancer subtype based on multi-omics data. *Molecular Therapy Nucleic Acids* 2025;**36**:102534.

Way GP, Greene CS. Extracting a biologically relevant latent space from cancer transcriptomes with variational autoencoders. *Pac Symp Biocomput* 2018;**23**:80-91.

Witten DM, Tibshirani RJ. Extensions of sparse canonical correlation analysis with applications to genomic data. *Stat Appl Genet Mol Biol* 2009;**8**:Article28.

Zhang Y, Kiryu H. MODEC: an unsupervised clustering method integrating omics data for identifying cancer subtypes. *Brief Bioinformatics* 2022;**23**:bbac372.

Zhao J, Zhao B, Song X *et al*. Subtype-DCC: decoupled contrastive clustering method for cancer subtype identification based on multi-omics data. *Brief Bioinformatics* 2023;**24**:bbad025.